\documentclass[10pt,twocolumn,letterpaper]{article}

\usepackage[pagenumbers]{cvpr} 

\usepackage{graphicx}
\usepackage{amsmath}
\usepackage{amssymb}
\usepackage{amsfonts}
\usepackage{booktabs}
\usepackage{color}
\usepackage[table]{xcolor}
\usepackage{multirow}
\usepackage{subcaption}
\usepackage{caption}
\usepackage[misc]{ifsym}
\usepackage[pagebackref,breaklinks=true,colorlinks,citecolor=blue,urlcolor=blue,linkcolor=blue,bookmarks=false]{hyperref}

\def\paperID{1763} 
\def\confName{CVPR}
\def\confYear{2024}

\title{Towards Language-Driven Video Inpainting \\ via Multimodal Large Language Models}

\author{
    Jianzong Wu$^{1,3}$\thanks{The work was done during the Shanghai AI Laboratory internship. \textsuperscript{$\dagger$}: Project Lead and corresponding author. Email: xiangtai94@gmail.com} \quad Xiangtai Li$^{2,3}$ \textsuperscript{$\dagger$} \quad Chenyang Si$^{2}$ \quad Shangchen Zhou$^{2}$ \quad Jingkang Yang$^{2}$ \quad Jiangning Zhang$^{5}$ \\ \quad Yining Li$^{3}$ \quad Kai Chen$^{3}$ \quad Yunhai Tong$^{1,4}$  \quad Ziwei Liu$^{2}$ \quad Chen Change Loy$^{2}$  \vspace{0.3em} \\
    {\normalsize $^1$ National Key Laboratory of General Artificial Intelligence, Peking University \quad $^2$S-Lab, Nanyang Technological University} \\
    {\normalsize  $^3$Shanghai AI Laboratory  \quad 
    $^4$ PKU-Wuhan Institute for Artificial Intelligence \quad $^5$ Zhejiang University  } \\
    {\normalsize Project Page: \url{https://jianzongwu.github.io/projects/rovi}}
}

\begin{document}
\maketitle

\begin{abstract}
We introduce a new task -- language-driven video inpainting, which uses natural language instructions to guide the inpainting process. This approach overcomes the limitations of traditional video inpainting methods that depend on manually labeled binary masks, a process often tedious and labor-intensive. We present the Remove Objects from Videos by Instructions (ROVI) dataset, containing 5,650 videos and 9,091 inpainting results, to support training and evaluation for this task. We also propose a novel diffusion-based language-driven video inpainting framework, the first end-to-end baseline for this task, integrating Multimodal Large Language Models to understand and execute complex language-based inpainting requests effectively. Our comprehensive results showcase the dataset's versatility and the model's effectiveness in various language-instructed inpainting scenarios. We have made datasets, code, and models publicly available at \url{https://github.com/jianzongwu/Language-Driven-Video-Inpainting}.
\vspace{-6mm}
\end{abstract}

\begin{figure}[t!]
	\centering
	\includegraphics[width=1.0\linewidth]{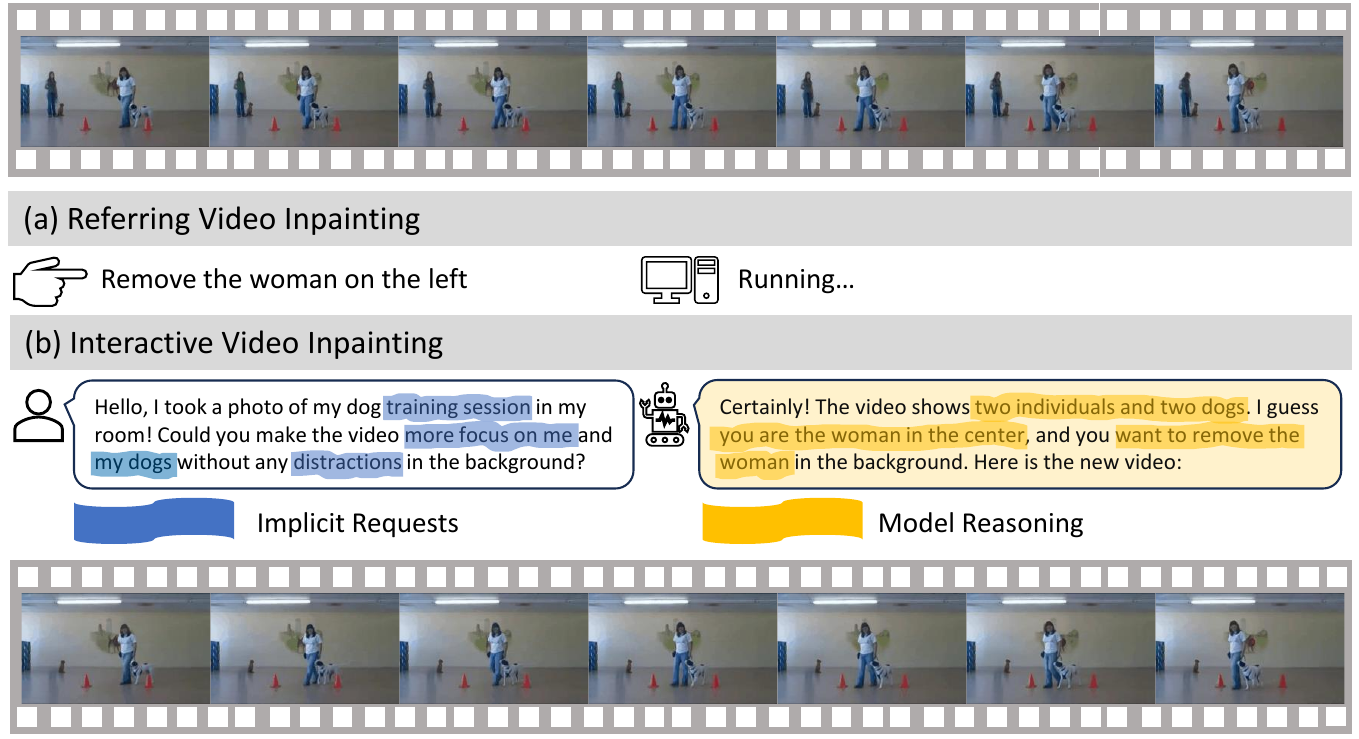}
	\caption{\small \textbf{Language-driven video inpainting.} It contains two sub-tasks based on the expression types. The referring video inpainting task takes simple referring expressions as input, while interactive video inpainting receives chat-style conversations. The conversation may encounter implicit requests, and the model needs to reason for a correct understanding.}
	\label{fig:teaser}
    \vspace{-5mm}
\end{figure}

\section{Introduction}

Video inpainting, a technique for restoring missing or corrupted segments in video frames, finds extensive application in areas such as video completion~\cite{video-inpaint-3d-gated-cnn}, video restoration~\cite{video-inpainting-video-restore}, and object removal~\cite{video-inpainting-object-removal}.
Despite continuous advancements in enhancing image quality and temporal coherence of inpainting results~\cite{video-inpainting-cnn-joint,video-inpainting-cnn-proposal,video-inpainting-end-to-end,propainter}, current methods predominantly depend on \textit{manually annotated binary masks} to identify restoration areas. 
This manual process is time-consuming and impractical for long videos. 
While automatic labeling tools, such as segmentation and object tracking models~\cite{inpaint-anything,ostrack,STTN}, offer some relief, they often necessitate manual refinement due to imperfect labeling.

Perhaps a more natural way to perform video inpainting is through natural language, as shown in~\cref{fig:teaser}. The task would become much easier if we could leverage natural language descriptions to specify the inpainting areas, like ``woman on the left,'' thereby preventing the need for pixel-level manual annotations.
Importantly, the language-driven setting can benefit from the flexibility of natural language. For example, with richer sentences, one can easily refer to multiple or abstract objects, which is much more effective than labeling masks.
Extending from this notion, one could divide the task into two subtasks, namely ``Referring Video Inpainting'' and ``Interactive Video Inpainting.'' The former takes simple referring expressions as inputs, and the latter considers more complex conversation-like interactions to accomplish the inpainting task.
%


\begin{figure}[t!]
	\centering
	\includegraphics[width=1.0\linewidth]{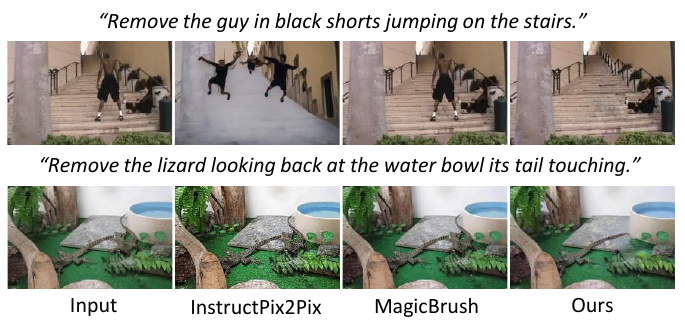}
	\caption{\small \textbf{Comparison with general image editing models}. InstructPix2Pix~\cite{instructpix2pix} and MagicBrush~\cite{magicbrush} are general image editing methods based on diffusion models. They produce inferior results when instructed to remove objects in videos.}
	\label{fig:general-editing-not-work}
    \vspace{-5mm}
\end{figure}

To establish a model for the proposed tasks, it is essential to have an appropriate dataset for both training and evaluation. Currently, no publicly available dataset comprises the triplets of original videos, removal expressions, and inpainted videos.
In response to this gap, we build a new dataset named the Remove Objects from Videos by Instructions (ROVI) dataset. 
Specifically, we employ referring object segmentation datasets, which are pre-annotated with object masks and descriptive expressions. These datasets are further augmented with corresponding inpainted videos generated using a state-of-the-art video inpainting model. 
However, we find existing referring expressions for interactive video inpainting tasks too simplistic.
To address this limitation, we employ Multimodal Large Language Models (MLLMs)~\cite{mllm-context-det,minigpt,gpt4-early-experiments} to create conversation-like dialogues. These dialogues are designed to simulate real-world scenarios, encompassing user requests and corresponding machine responses. This approach enriches the dataset, making it more representative of the complexity and variability found in practical video inpainting applications.

In addition to the dataset, we introduce the first end-to-end baseline model, Language-Driven Video Inpainting (LGVI), for the proposed tasks. Our model is built upon diffusion-based generative models. 
In particular, we inflate the text-to-image (T2I) model to become a text-to-video (T2V) architecture by extending the parameters with an additional temporal dimension. We propose an efficient visual conditioning approach that minimally increases the number of parameters. To further enhance our model's capabilities for the interactive task, we extend the LGVI framework to LGVI-I (Interactive). This extension incorporates an MLLM specifically designed to process and understand user requests phrased in a conversation-like format. The LGVI-I model is trained in an end-to-end manner. This interactive architecture enables the system to interpret complex instructions accurately. As a result,  it can produce appropriate inpainting results and relevant responses within the conversational context, thus paving the way for more intuitive and flexible user interactions with interactive video inpainting systems.

In summary, our key contributions are as follows: 
\begin{itemize}
    \item We introduce a novel language-driven video inpainting task, significantly reducing reliance on human-labeled masks in video inpainting applications. This task includes two distinct sub-tasks: referring video inpainting and interactive video inpainting.
    \item We propose a dataset to facilitate training and evaluation for the proposed tasks. This dataset is the first of its kind, containing triplets of original videos, removal expressions, and inpainted videos, offering a unique resource for research in this domain.
    \item We present a diffusion-based architecture, LGVI, as a baseline model for the proposed task. We show how one could leverage MLLMs to improve language guidance for interactive video inpainting. To our knowledge, it is the first model to perform end-to-end language-driven video inpainting.
\end{itemize}

\begin{table*}[t]
    \centering
    \caption{\small \textbf{Comparison between the ROVI dataset and related datasets.} We choose two commonly used image inpainting (II) datasets, two video inpainting (VI) datasets, and one language-guided image inpainting (LII) dataset. Our ROVI dataset is the first for language-guided video inpainting (LVI) and interactive video inpainting (IVI) tasks.}
    \label{tab:dataset-statistics-compare}
    \scalebox{0.585}{
    \hspace{-2mm}
    \begin{tabular}{lccccccccccccc}
\toprule[0.15em]
\multirow{2}{*}{Dataset} & \multirow{2}{*}{Task} & \multirow{2}{*}{Scene} & \multirow{2}{*}{\#Images} & \multirow{2}{*}{\#Videos} & \multirow{2}{*}{\#Frames} & \multicolumn{4}{c}{Annotations} & \multirow{2}{*}{\#Objects} & \multirow{2}{*}{\#Exprs} & \multirow{2}{*}{\#Chats} & \multirow{2}{*}{Source} \\
& & & & & & mask & expr & inpaint & chat & & & & \\
\hline
Places~\cite{Places} & II & Buildings \& Places & 10,624,928 & - & - & $\times$ & $\times$ & $\times$ & $\times$ & - & - & - & - \\ 
CelebA~\cite{celebA} & II & Human Face & 202,599 & - & - & $\times$ & $\times$ & $\times$ & $\times$ & - & - & - & - \\
\hline
YouTube-VOS~\cite{xu2018youtube} & VI & General & - & 4,453 & 197,272 & $\checkmark$ & $\times$ & $\times$ & $\times$ & 7,755 & - & - & YouTube \\
DAVIS~\cite{davis} & VI & General & - & 50 & 3,455 & $\times$ & $\times$ & $\times$ & $\times$ & 3,455 & - & - & - \\
\hline
GQA-Inpaint~\cite{inst-inpaint} & LII & General & 49,311 & - & - & $\checkmark$ & $\checkmark$ & $\checkmark$ & $\times$ & 97,854 & 107,252 & - & GQA~\cite{gqa} \\
\hline
ROVI & LVI + IVI & General & - & 5,650 & 247,018 & $\checkmark$ & $\checkmark$ & $\checkmark$ & $\checkmark$ & 9,091 & 12,534 & 9,091 & Refer-YouTube-VOS~\cite{refer-youtube-vos} + A2D-Sentences~\cite{a2d-sentences} \\
\bottomrule[0.15em]
     \end{tabular}}
\end{table*}

\begin{figure*}[t!]
    \centering
    \begin{subfigure}{0.21\textwidth}
      \includegraphics[width=\linewidth]{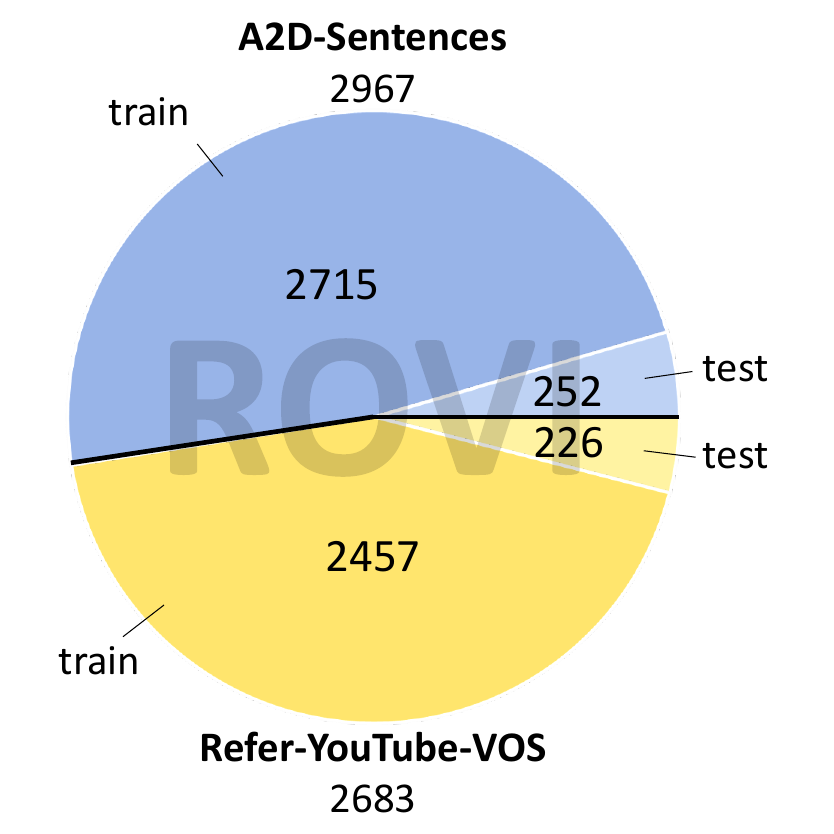}
      \caption{Number of videos and split.}
      \label{fig:dataset-statistics-video-source}
    \end{subfigure}\hfill
    \begin{subfigure}{0.23\textwidth}
      \includegraphics[width=\linewidth]{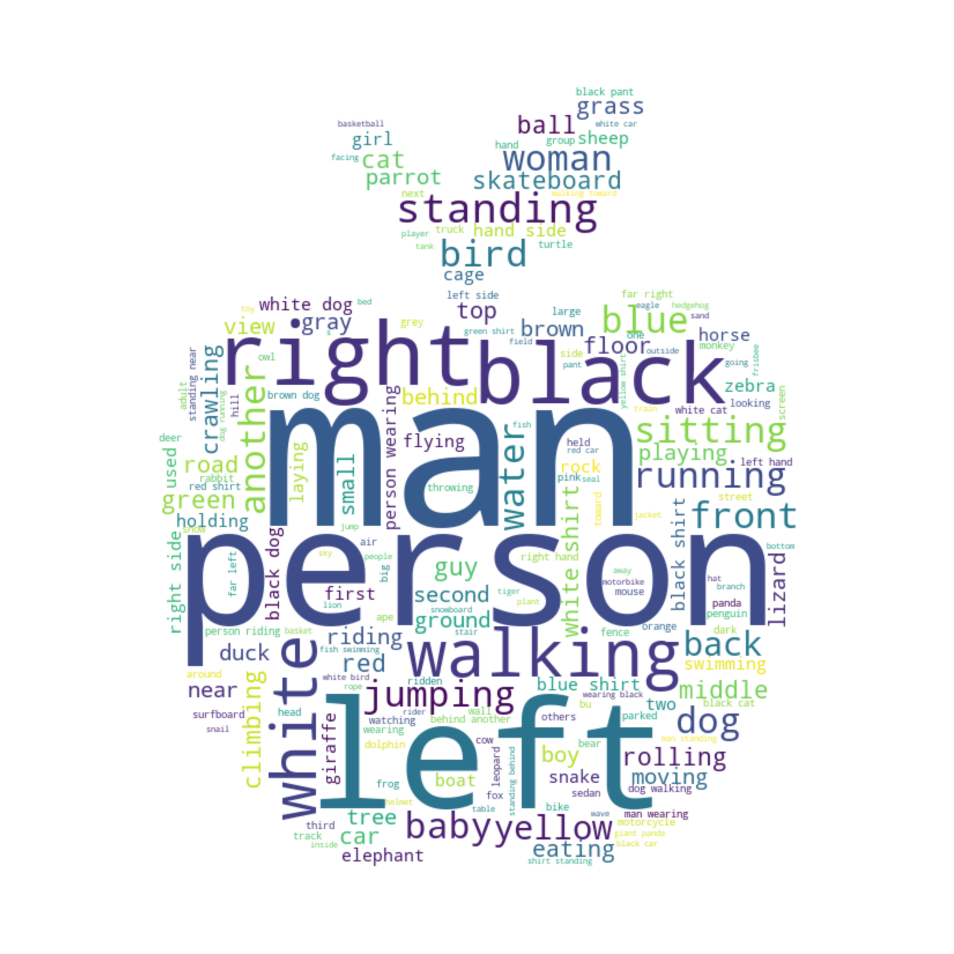}
      \caption{Word cloud of expressions.}
      \label{fig:dataset-statistics-cloud-apple}
    \end{subfigure}\hfill
    \begin{subfigure}{0.54\textwidth}
      \includegraphics[width=\linewidth]{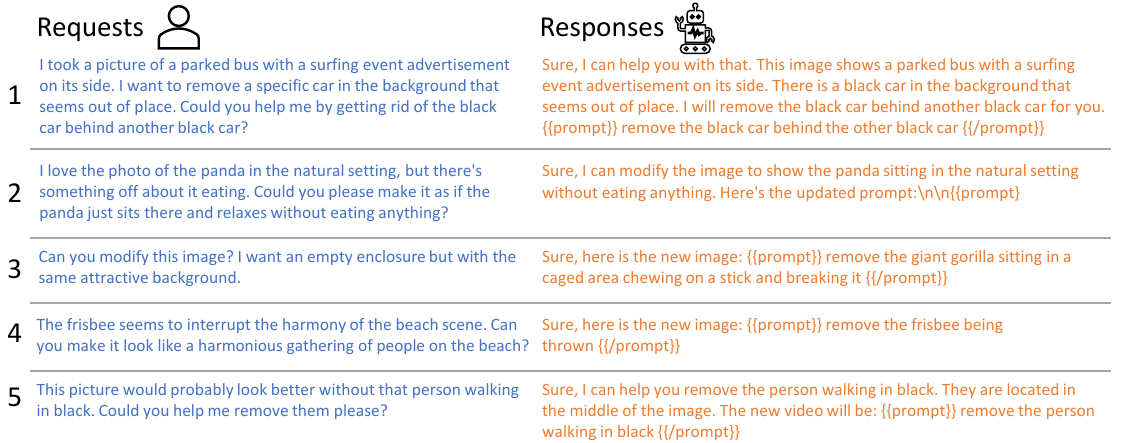}
      \caption{Examples of interactive requests and responses.}
      \label{fig:dataset-statistics-expr-examples}
    \end{subfigure}
    
    
    \begin{subfigure}{0.27\textwidth}
      \includegraphics[width=\linewidth]{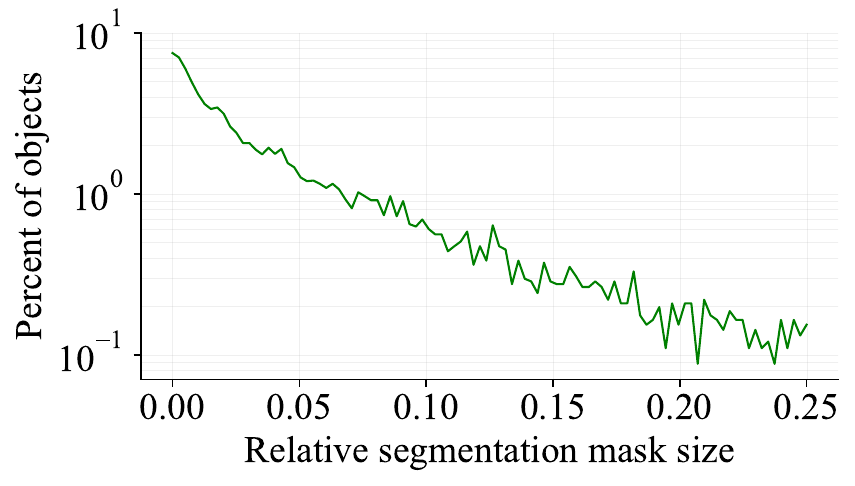}
      \caption{Relative segmentation mask size.}
      \label{fig:dataset-statistics-mask-area-distribution}
    \end{subfigure}\hfill
    \begin{subfigure}{0.27\textwidth}
      \includegraphics[width=\linewidth]{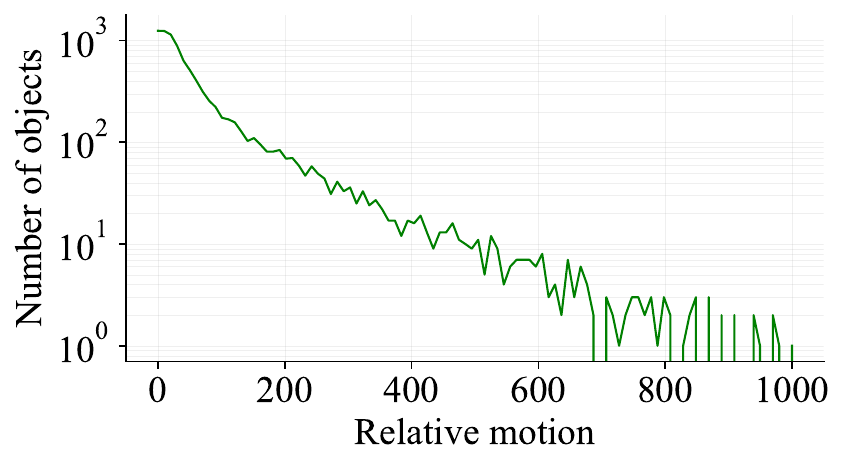}
      \caption{Relative object motion distribution.}
      \label{fig:dataset-statistics-motion-distribuion}
    \end{subfigure}\hfill
    \begin{subfigure}{0.45\textwidth}
      \includegraphics[width=\linewidth]{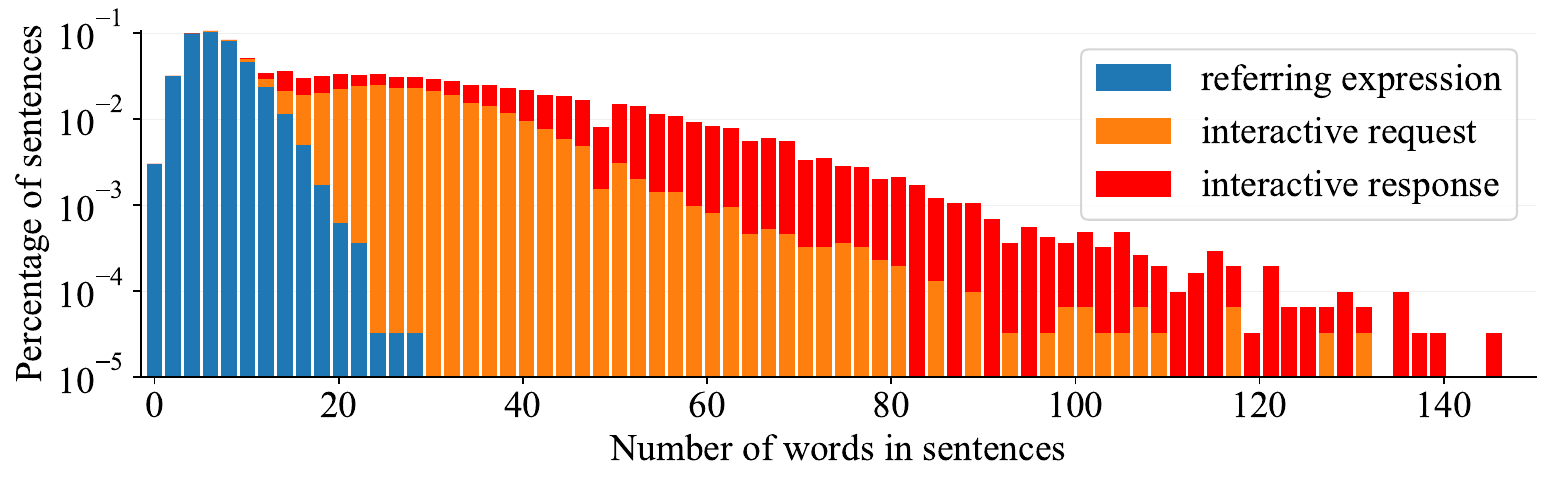}
      \caption{Sentence lengths distribution.}
      \label{fig:dataset-statistics-sentence-distribution}
    \end{subfigure}
    
    \caption{\textbf{The ROVI dataset statistics.} Best viewed in color.}
    \vspace{-5mm}
    \label{fig:dataset-statistics}
\end{figure*}

\section{Related Work}



\noindent
\textbf{Video inpainting.}
Video inpainting is a technique aimed at restoring or filling missing or corrupted parts in a video plausibly. 
While related to image inpainting methods~\cite{image-inpainting-context,image-inpainting-gated-cnn,image-inpainting-holes,image-inpainting-partial,image-inpainting-recurrent,image-inpainting-transformer-hole}, video inpainting techniques~\cite{liu2021fuseformer,video-inpaint-3d-gated-cnn,video-inpainting-cnn-joint,video-inpainting-cnn-proposal,video-inpainting-end-to-end,video-inpainting-flow-guided,video-inpainting-short-long,propainter} extend the problem to the more complex domain of moving visuals. 
This technique can be applied for various applications, such as object removal, visual restoration, and completion. With the advent of deep learning, visual inpainting networks usually employ convolutional neural networks (CNNs)~\cite{image-inpainting-gated-cnn,video-inpainting-cnn-joint,video-inpainting-cnn-proposal} and generative adversarial networks (GANs)~\cite{image-inpainting-context,image-inpainting-holes,image-inpainting-partial,image-inpainting-recurrent}. Recent works also apply vision Transformers~\cite{ViT,vit-token-1,vit-token-2,vit-token-3,liu2021swin} to enhance the global interaction among visual features~\cite{image-inpainting-transformer-hole,video-inpainting-end-to-end,video-inpainting-flow-guided,video-inpainting-short-long,propainter}. State-of-the-art methods show strong abilities in restoring missing parts and removing objects.
Most of these works require the input of a binary mask to define the restoring area~\cite{video-inpainting-end-to-end,image-inpainting-transformer-hole,image-inpainting-holes,liu2021fuseformer,video-inpainting-flow-guided,video-inpaint-3d-gated-cnn}. However, the generation of object-like masks, particularly for lengthy videos, poses a significant and labor-intensive challenge,


\noindent
\textbf{Language-driven visual editing.}
Diffusion-based text-to-image generation models (DMs)~\cite{stable-diffusion,diffusion-glide,diffusion-imagen,diffusion-unclip,make-a-video} show excellent abilities in generating images and videos following text guidance. Recent studies also achieve image editing~\cite{prompt2prompt,instructpix2pix,sdedit,image-edit-fidelity,magicbrush}, image segmentation and grouping~\cite{zhou2023edgesam,wu2022towards,wu2023towards,wu2023betrayed,li2023transformer,li2024omg,xu2024rapsam,yuan2024ovsam,sfnet} and video editing~\cite{tune-a-video,text2video-zero} with DMs.
Among these, Prompt2Prompt~\cite{prompt2prompt} manipulates the cross-attention maps within the model to enable various editing operations such as object modification, addition, and style transfer. InstructPix2Pix~\cite{instructpix2pix} leverages this approach to create an image editing dataset. Similarly, Tune-A-Video~\cite{tune-a-video} proposes a training-free architecture to edit videos by language references.
However, these works are intended for general visual editing. They tend to yield suboptimal results when applied to more specific challenges, such as language-driven video inpainting. Figure~\ref{fig:general-editing-not-work} shows two examples where these models produce inferior results when instructed to remove objects.
A few works have explored the image inpainting task using DMs. Repaint~\cite{repaint} takes the image and mask as input and lets the DM restore the original image. SmartBrush~\cite{smartbrush} takes mask and text as input to guide a region-controlled generation on the masked area, which aims to generate new concepts rather than remove the object. 
Recently, Inst-Inpaint~\cite{inst-inpaint} proposes a method to perform object removal on images based on the language descriptions. Despite its innovative approach, Inst-Inpaint's training samples are constrained by a lack of interactive expressions and video resources, which limits its practical effectiveness in complex scenarios.


\noindent
\textbf{Multi-Modal Large Language Models.} Large language models (LLMs) have demonstrated exceptional performance across a variety of text-based tasks and applications~\cite{gpt4-early-experiments,GPT-3,openai2023gpt4,llama,mimic-it,planting-a-seed-of-vision}. Recent works extend the capabilities of LLMs to include image processing and computer vision. A notable example is LLaVA~\cite{llava}, which translates image tokens into a language feature space, thereby transforming the fine-tuned model into an MLLM. This adaptation enables MLLMs to interpret and understand visual content.  Subsequent research uses MLLMs in diverse applications, including visual reasoning, object detection, and segmentation~\cite{llava,minigpt,detgpt,mllm-context-det,mimic-it,planting-a-seed-of-vision}. To the best of our knowledge, this study is the first to integrate MLLMs into the domain of language-driven video inpainting. 
\begin{figure*}[!t]
	\centering
	\includegraphics[width=1\linewidth]{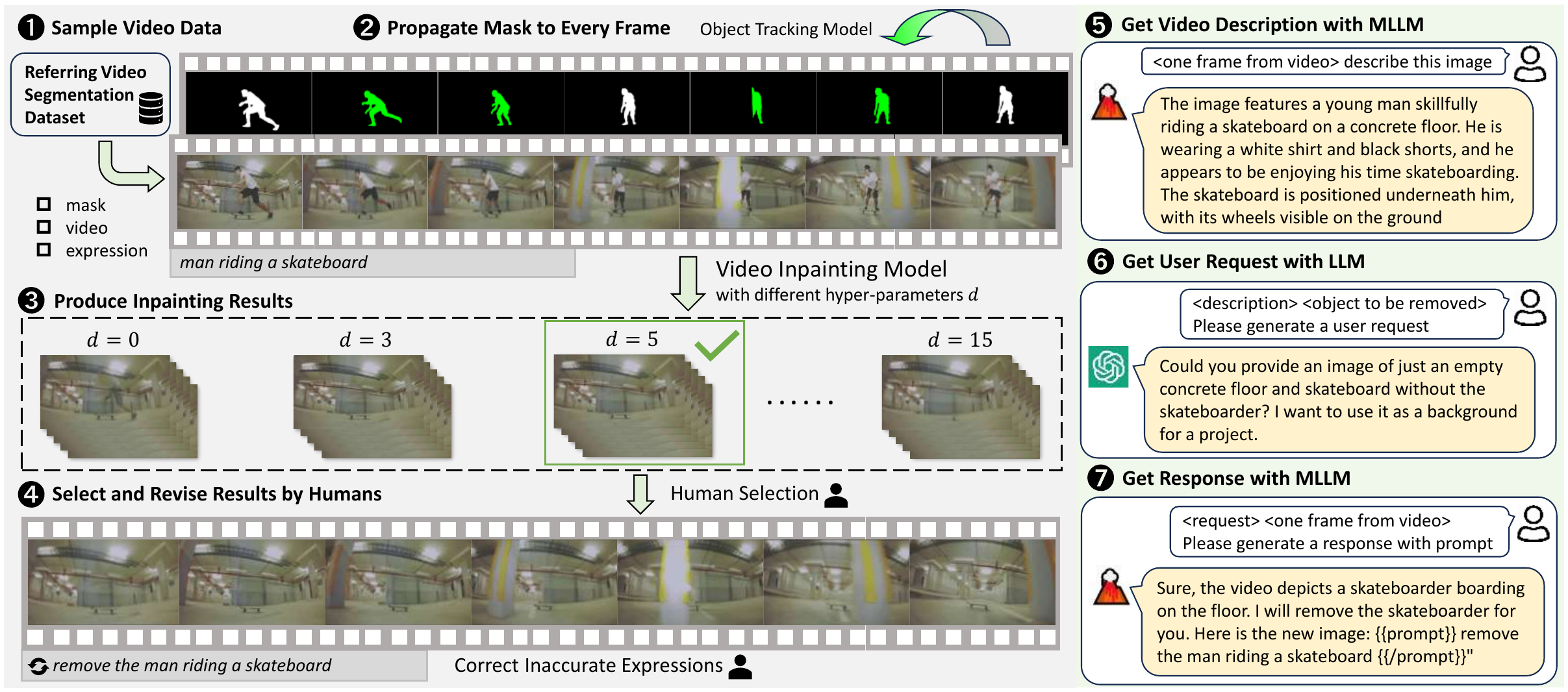}
	\caption{\small \textbf{ROVI dataset annotation pipeline}. The building process of the ROVI dataset involves two distinct phases: inpainting annotation and interactive annotation. In the inpainting annotation phase, the primary objective is to incorporate inpainting results into existing referring video segmentation datasets, which initially contain object masks and expressions. During the interactive annotation pipeline, we follow a multi-step approach incorporating LLMs and MLLMs. Best viewed in color.}
	\label{fig:dataset-rovi-pipeline}
    \vspace{-5mm}
\end{figure*}

\section{ROVI Dataset}

\subsection{Comparing with Existing Datasets}

Table~\ref{tab:dataset-statistics-compare} summarizes the differences between ROVI and related datasets.
In image and video inpainting research, prevalent training and evaluation datasets mainly include vision-centric ones like Places~\cite{Places}, CelebA~\cite{celebA}, YouTube-VOS~\cite{xu2018youtube}, and DAVIS~\cite{davis}, without human annotations. 
These datasets typically employ random masking in training to simulate missing areas for inpainting. However, for object removal tasks, specifically labeled masks are essential. While YouTube-VOS provides object masks, it lacks corresponding inpainting ground truths. The GQA-Inpaint dataset~\cite{inst-inpaint}, although rich in object expressions and inpainting results, is limited to image data and does not accommodate video or interactive contexts. 
Our ROVI dataset addresses these limitations with comprehensive annotations covering a wide array of regions, including object masks, referring expressions, inpainting results, and conversation-like dialogues. Unlike Places and CelebA, which focus on specific image categories like buildings or faces, ROVI encompasses a broader spectrum of general scenes, making it more adaptable for diverse inpainting applications.

\subsection{Dataset Statistics}

Figure~\ref{fig:dataset-statistics} presents a comprehensive statistical analysis of the ROVI dataset. The dataset encompasses 2,967 videos from A2D-Sentences and 2,683 videos from Refer-YouTube-VOS, divided into train and test splits, as shown in \cref{fig:dataset-statistics-video-source}. Figure~\ref{fig:dataset-statistics-cloud-apple} illustrates the diversity of referring expressions with word clouds. Figure~\ref{fig:dataset-statistics-expr-examples} shows several examples of our dataset's interactive requests and responses, showing the diversity and complexity of the dialogues. Figure~\ref{fig:dataset-statistics-mask-area-distribution} details the relative sizes of segmentation masks (mask area divided by image area). We drop objects with a relative size larger than 0.25 because the inpainting results for large objects usually have worse qualities. Figure~\ref{fig:dataset-statistics-motion-distribuion} analyzes the distribution of object motion. Figure~\ref{fig:dataset-statistics-sentence-distribution} delivers a histogram of sentence lengths within the dataset.

\subsection{Dataset Construction Pipeline}


\noindent
\textbf{Video data selection.}
As depicted in \cref{fig:dataset-rovi-pipeline}, we have chosen referring video object segmentation datasets for the source of video data. Referring video object segmentation aims to segment an object referred to by a given language description. These datasets have pre-annotated object masks and descriptive expressions, making them well-suited for the proposed task. Specifically, we select Refer-YouTube-VOS~\cite{refer-youtube-vos} and A2D-Sentences~\cite{a2d-sentences} as our data sources.
%
%

\noindent
\textbf{Annotation pipeline.}
We use a video inpainting model to generate the inpainting ground truth. Specifically, we choose E$^2$FGVI~\cite{video-inpainting-end-to-end}, a state-of-the-art video inpainting model, to produce the inpainting results. This model, trained on video data, guarantees temporal consistency in the inpainting results. 

To further ensure the ground truth is of high quality, we incorporate a human selection process on the hyperparameter of the inpainting method. Specifically, the input mask can be expanded with different pixel sizes, denoted as $d$. The bigger the $d$ is, the larger the input mask is developed so that it may cover the whole object. The best $d$ value varies through objects, causing an unstable performance in the inpainted videos if set to a fixed value. Therefore, throughout the data generation process, we experiment with various hyperparameters to generate multiple results for each object and involve human annotators to select the best result. More details are provided in the supplementary.

%
%
For interactive annotations, we need to collect expressions through chat-style conversations. Unlike the straightforward ``remove'' sentences, these interactive requests should be implicit, necessitating the model to discern the user's underlying intent.
Rather than relying solely on human annotators to articulate these requests, we explore a more automated approach: we employ LLMs and MLLMs to simulate a human user and generate potential requests and responses. We propose a multi-step approach with details illustrated in~\cref{fig:dataset-rovi-pipeline}.
%
%
%
By employing this dual-faceted annotation pipeline, the ROVI dataset is enabled to handle complex user requests. 

\begin{figure*}[t!]
	\centering
	\includegraphics[width=0.90\linewidth]{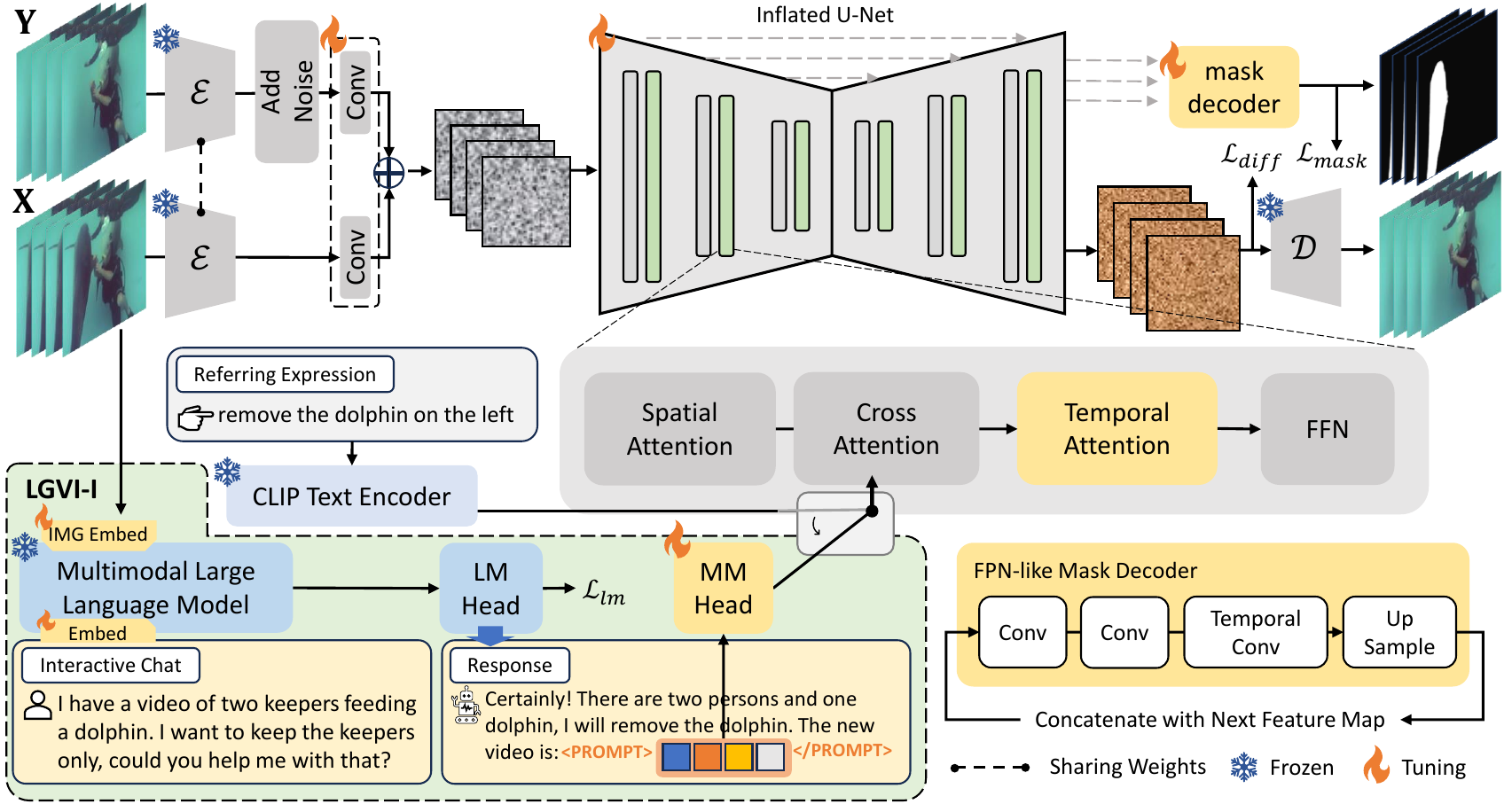}
	\caption{\small \textbf{The training process of LGVI and LGVI-I}. We inflate the U-Net with a temporal dimension to allow video input. To ensure temporal consistency in the generated videos, we introduce a temporal attention module between cross-attention and FFN layers. Additionally, we propose a mask decoder module for explicit guidance in inpainting tasks. We augment LGVI with MLLM joint training for interactive video inpainting, resulting in LGVI-I as the baseline. The output of MLLM includes a set of prompt tokens, which is fed into the cross attention of the U-Net.}
	\label{fig:model-LGVI}
\end{figure*}

\section{Methodology}

In this section, we introduce our Language-Driven Video Inpainting framework (LGVI) and the MLLM-enhanced LGVI-I (Interactive) architecture. The latter is built from the former architecture by adding extra LLM as language controllers.

\subsection{LGVI}

The LGVI framework is shown in~\cref{fig:model-LGVI}, which is built on the architecture of Stable Diffusion~\cite{stable-diffusion}. To extend the framework to video inputs, we perform temporal inflation by reorganizing the network's structure following~\cite{tune-a-video, make-a-video}. For a batch video input with $T$ frames, denoted as $\mathbf{X} \in \mathbb{R}^{B\times T\times H\times W\times 3}$, where $B$ is the batch size, and $H\times W$ are the size, we transpose the tensor to $\mathbf{X} \in \mathbb{R}^{(B\times T)\times H\times W\times 3}$. This transformation converts the input into a 4-dimensional image batch input format. The pre-trained 2D networks can process video clips as they are separate images. Additionally, we introduce a parameter-efficient Temporal Attention module positioned between the cross-attention and FFN network. Given latent feature $\mathbf{v} \in \mathbb{R}^{(B\times T)\times D\times C}$, where $D$ is the length of patched visual tokens, and $C$ is the channel size, we transpose it to $\mathbf{v}' \in \mathbb{R}^{(B\times D)\times T\times C}$. The Temporal Attention module is formulated as follows:
\begin{equation}
    \begin{aligned}
        \mathbf{Q} = \mathbf{W}_q&\mathbf{v}, \quad \mathbf{K} = \mathbf{W}_k\mathbf{v}, \quad \mathbf{V} = \mathbf{W}_v\mathbf{v}, \\
        \mathrm{Attention}&(\mathbf{Q},\mathbf{K},\mathbf{V}) = \mathrm{Softmax}(\frac{\mathbf{Q}\mathbf{K}^T}{\sqrt{C}}) \cdot \mathbf{V},
    \end{aligned}
\end{equation}
where $\mathbf{W}_q$, $\mathbf{W}_k$, and $\mathbf{W}_v$ are learnable matrices to project the inputs to query, key, and value.
%
%
The computational complexity of the Temporal Attention module is $\mathcal{O}(CT^2)$, while spatial self-attention has a complexity of $\mathcal{O}(CD^2)$. Considering $T \ll D$. The Temporal Attention module is a time-efficient tool to ensure the consistency of video sequences. 

Diffusion models learn to gradually remove noises in a noised video. During training, the target video $\mathbf{Y}$ is added with noises to be the start point of the noised video. Besides the noised target video input, LGVI also takes the original video $\mathbf{X}$ as a control signal input. Concretely, we encode the original video $\mathbf{X}$ to the latent space and concatenate its feature with the noised target video in the channel dimension. Note that the noise is added only to the target video latent, and during testing, the noised target video is a randomly generated noise.
\begin{equation}
    \begin{aligned}
        \mathbf{z}_0 &= \mathcal{E}(\mathbf{Y}), \quad \mathbf{c}_x = \mathcal{E}(\mathbf{X}), \\
        \mathbf{z}_t &= \mathrm{AddNoise}(\mathbf{z}_0), \\
        \mathbf{v}_t &= \mathrm{Conv}_v(\mathbf{z}_t) + \mathrm{Conv}_x(\mathbf{c}_x), \\
    \end{aligned}
\end{equation}
where $\mathcal{E}$ is the pre-trained VAE encoder, $t$ is the sampled timestamp, $\mathrm{Conv}_v$ and $\mathrm{Conv}_x$ are convolutional layers with $3\times 3$ kernels to transfer the latent codes into U-Net feature dimensions. The initial weights of $\mathrm{Conv}_x$ are set to all-zero. This technique allows the model to add video condition guidance during training.
Due to mask annotations in the ROVI dataset, we can leverage masks as an additional supervision signal in our LGVI framework. Concretely, we implement a mask decoder to predict the object's mask in the video that needs to be inpainted or removed. This decoder uses the outputs from U-Net up-blocks and consists of convolutional and temporal convolutional layers. The use of mask supervision enables the model to focus on the region described in the natural language input, thereby facilitating precise and targeted inpainting. The effectiveness of mask supervision can be seen in~\cref{sec:exp}. The training objective of LGVI is:
\begin{equation}
    \begin{aligned}
        \mathcal{L}_{diff} &= \mathbb{E}_{\mathbf{X}, \mathbf{\epsilon} \sim \mathcal{N}(\mathbf{0}, \mathbf{I}), t} \left [ || \mathbf{\epsilon} - \mathbf{\epsilon}_\theta(\mathbf{v}_t, \mathbf{c}, t) ||^2_2 \right ], \\
        \mathcal{L}_{mask} &= \mathrm{CrossEntropyLoss}(\hat{\mathbf{M}}, \mathbf{M}), \\
        \mathcal{L} &= \lambda_1 \mathcal{L}_{diff} + \lambda_2 \mathcal{L}_{mask},
    \end{aligned}
\end{equation}
where $\mathcal{L}_{diff}$ and $\mathcal{L}_{mask}$ are the diffusion model training objective and mask loss, respectively; $\mathbf{c}$ is the language guidance features from the referring expressions; $\hat{\mathbf{M}}$ is the mask prediction and $\mathbf{M}$ is the ground truth mask. The parameters $\lambda_1$ and $\lambda_2$ are loss weights to balance training.

\subsection{LGVI-I with MLLM}
In the interactive video inpainting task, models are expected to extract valuable information from complex conversations. To overcome this problem, we propose incorporating MLLMs to extend the LGVI from work to LGVI-I (Interactive). MLLMs have demonstrated strong capabilities in visual comprehension and multimodal reasoning, making them well-suited for our proposed interactive video inpainting task. As shown in~\cref{fig:model-LGVI}, the MLLM takes both the image frame and the chat-style user request as inputs, generating the language response and a pair of special indicators: $<$PROMPT$>$ and $<$/PROMPT$>$. The hidden states of the last layer between these two indicators are then passed through an MM head, implemented as a two-layer linear layer with activation functions. The transformed features are fed to the cross-attention module to guide the U-Net inpainting process. Mathematically, given the input video $\mathbf{X}$ and user request $s$, the computation pipeline of the MLLM can be summarized as follows:
\begin{equation}
    \begin{aligned}
        \mathbf{e}_l &= f(s), \quad \mathbf{e}_i = \mathbf{W}_{trans} \cdot g(\mathbf{X}_0), \\
        \mathbf{h} &= \mathrm{MLLM}([\mathbf{e}_l, \mathbf{e}_v]), \\
        \hat{\mathbf{w}} &= \mathbf{W}_{lm} \cdot \mathbf{h}, \\
        \mathbf{h}_p &= \mathbf{W}_{mm} \cdot \mathrm{find\_prompt}(\hat{\mathbf{w}}, \mathbf{h}), \\
    \end{aligned}
\end{equation}
where $f$ is the language token embedding and $g$ is a pre-trained image backbone to extract image features. $\mathbf{W}_{trans}$ is a linear layer that transforms image features into language token space. $\mathbf{h}$ is the last layer's hidden states of the MLLM. $\hat{\mathbf{w}}$ is the predicted language token distribution through the LM head. $\mathbf{W}_{lm}$ is the weights of the LM head. Among the predicted words, we use $\mathrm{find\_prompt}$ function to find the $<$PROMPT$>$ and $<$/PROMPT$>$ indicator and extract the hidden states that lie between these two indicators. $\mathbf{W}_{lm}$ is the weights of MM head. The MM head transfers the selected tokens into $\mathrm{h}_p$, which is then fed into the U-Net cross-attention module. In this process, $\mathrm{h}_p$ serves as vision-aware language guidance for the inpainting process.
The training objective of LGVI-I is:
\begin{equation}
    \begin{aligned}
        \mathcal{L}_{diff} &= \mathbb{E}_{\mathbf{X}, \mathbf{\epsilon} \sim \mathcal{N}(\mathbf{0}, \mathbf{I}), t} \left [ || \mathbf{\epsilon} - \mathbf{\epsilon}_\theta(\mathbf{v}_t, \mathbf{h}_p, t) ||^2_2 \right ], \\
        \mathcal{L}_{lm} &= \mathrm{CrossEntropyLoss}(\hat{\mathbf{w}}, \mathbf{w}), \\
        \mathcal{L} &= \lambda_1 \mathcal{L}_{diff} + \lambda_2 \mathcal{L}_{mask} + \lambda_3 \mathcal{L}_{lm},
    \end{aligned}
\end{equation}
where $\mathbf{h}_p$ is the MLLM-enhanced language condition, $\mathcal{L}_{lm}$ is language modeling loss, implemented as the Cross-Entropy Loss, $\mathbf{w}$ is the ground truth sentence, and $\lambda_3$ is the loss weight for language modeling loss.
By integrating an MLLM into the LGVI framework, the system achieves a higher level of user interactivity. This enables users to guide the visual inpainting process with interactive language instructions, thus establishing a more user-friendly and accessible approach for the interactive video inpainting task.

\begin{table}[!t]
    \centering
    \caption{\small \textbf{Quantitative results on the referring video inpainting task.} $E^*_{warp}$ denotes $E_{warp}(\times 10^{-2})$.}
    \label{tab:exp-refer-exprl}
    \scalebox{0.84}{
    \hspace{-1.7mm}
    \begin{tabular}{l|cccc}
\toprule[0.15em]
Method & PSNR $\uparrow$ & SSIM $\uparrow$ & VFID $\downarrow$ & $E^*_{warp}$ $\downarrow$ \\
\hline 
\multicolumn{5}{c}{\cellcolor{gray!25}Image Models} \\
\hline
InstructPix2Pix~\cite{instructpix2pix} & 18.12 & 0.600 & 0.361 & 1.343 \\
Inst-Inpaint~\cite{inst-inpaint} & 19.00 & 0.896 & 0.310 & 1.206 \\
MagicBrush~\cite{magicbrush} & 20.39 & 0.725 & 0.310 & 0.934 \\
\hline
\multicolumn{5}{c}{\cellcolor{gray!25}Multi-Stage Video Model} \\
\hline
Inpaint Anything*~\cite{inpaint-anything} & 22.84 & 0.728 & \textbf{0.283} & 0.874 \\
\hline
\multicolumn{5}{c}{\cellcolor{gray!25}One-Stage Video Model} \\
\hline
LGVI (Ours) & \textbf{22.85} & \textbf{0.756} & 0.308 & \textbf{0.901} \\
\bottomrule[0.15em]
     \end{tabular}}
\end{table}

\begin{table}[!t]
    \vspace{-2mm}
    \centering
    \caption{\small \textbf{Results on interactive video inpainting task.} $E^*_{warp}$ denotes $E_{warp}(\times 10^{-2})$. MB represents MagicBrush, and IA* represents Inpaint Anything*. The small numbers on the top 5 rows are compared with the referring video inpainting results. The small numbers on the last row are compared with LGVI.}
    \label{tab:exp-chat-exprl}
    \scalebox{0.73}{
    \hspace{-2mm}
    \begin{tabular}{l|cccc}
\toprule[0.15em]
Method & PSNR $\uparrow$ & SSIM $\uparrow$ & VFID $\downarrow$ & $E^*_{warp}$ $\downarrow$ \\
\hline 
\multicolumn{5}{c}{\cellcolor{gray!25}Image Models} \\
\hline
InstructPix2Pix~\cite{instructpix2pix} & 16.53{\color{red}\scriptsize (-1.59)} & 0.558{\color{red}\scriptsize (-0.042)} & 0.391{\color{red}\scriptsize (-0.003)} & 1.789{\color{red}\scriptsize (-0.446)} \\
Inst-Inpaint~\cite{inst-inpaint} & 18.96{\color{red}\scriptsize (-0.04)} & 0.702 & 0.314{\color{red}\scriptsize (-0.004)} & 1.047 \\
MagicBrush~\cite{magicbrush} & 20.46 & 0.728 & 0.311{\color{red}\scriptsize (-0.001)} & 0.901 \\
\hline
\multicolumn{5}{c}{\cellcolor{gray!25}Multi-Stage Video Model} \\
\hline
IA*~\cite{inpaint-anything} & 20.64{\color{red}\scriptsize (-2.20)} & 0.664{\color{red}\scriptsize (-0.064)} & 0.312{\color{red}\scriptsize (-0.029)} & 1.182{\color{red}\scriptsize (-0.308)} \\
\hline
\multicolumn{5}{c}{\cellcolor{gray!25}One-Stage Video Model} \\
\hline
LGVI (Ours) & 20.70{\color{red}\scriptsize (-2.15)} & 0.707{\color{red}\scriptsize (-0.049)} & 0.332{\color{red}\scriptsize (-0.024)} & 1.191{\color{red}\scriptsize (-0.290)} \\
\hline
\hline
\multicolumn{5}{c}{\cellcolor{gray!25}MLLM-Enhanced Two-Stage Model} \\
\hline
MB + MLLM & 20.37 & 0.726 & 0.313 & 1.004 \\
IA* + MLLM & 21.37 & 0.722 & 0.291 & 0.875 \\
LGVI + MLLM  & 21.45 & \textbf{0.738} & 0.311 & 0.923 \\
\hline
\multicolumn{5}{c}{\cellcolor{gray!25}MLLM-Enhanced End-to-End Model} \\
\hline
LGVI-I (Ours) & \textbf{22.24{\color{orange}\scriptsize (+1.54)}} & 0.732{\color{orange}\scriptsize (+0.025)} & \textbf{0.299{\color{orange}\scriptsize (+0.033)}} & \textbf{0.867{\color{orange}\scriptsize (+0.324)}} \\
\bottomrule[0.15em]
     \end{tabular}}     
\end{table}

\begin{figure}[t!]
	\centering
	\includegraphics[width=1.0\linewidth]{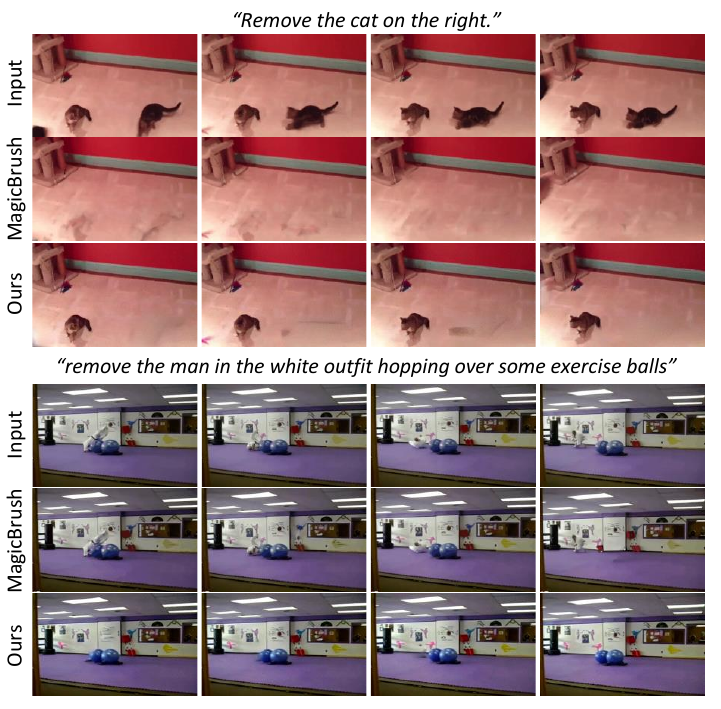}
	\caption{\small Qualitative comparison between LGVI and MagicBrush~\cite{magicbrush} on the referring video inpainting task.}
	\label{fig:qualitative-comparison-referring}
\end{figure}

\begin{figure}[t!]
	\centering
	\includegraphics[width=1.0\linewidth]{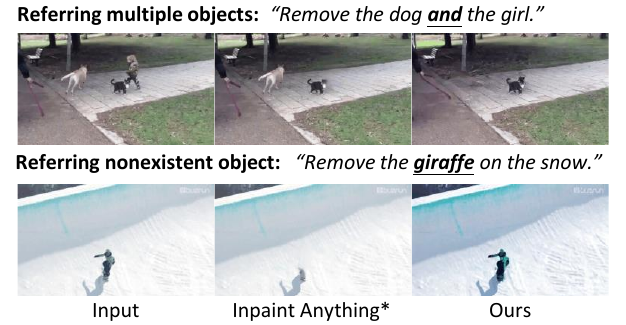}
	\caption{\small Examples of referring to multiple objects in one sentence and referring to nonexistence objects.}
	\label{fig:qualitative-multiple-nonexist}
\end{figure}

\begin{figure*}[t!]
	\centering
	\includegraphics[width=0.85\linewidth]{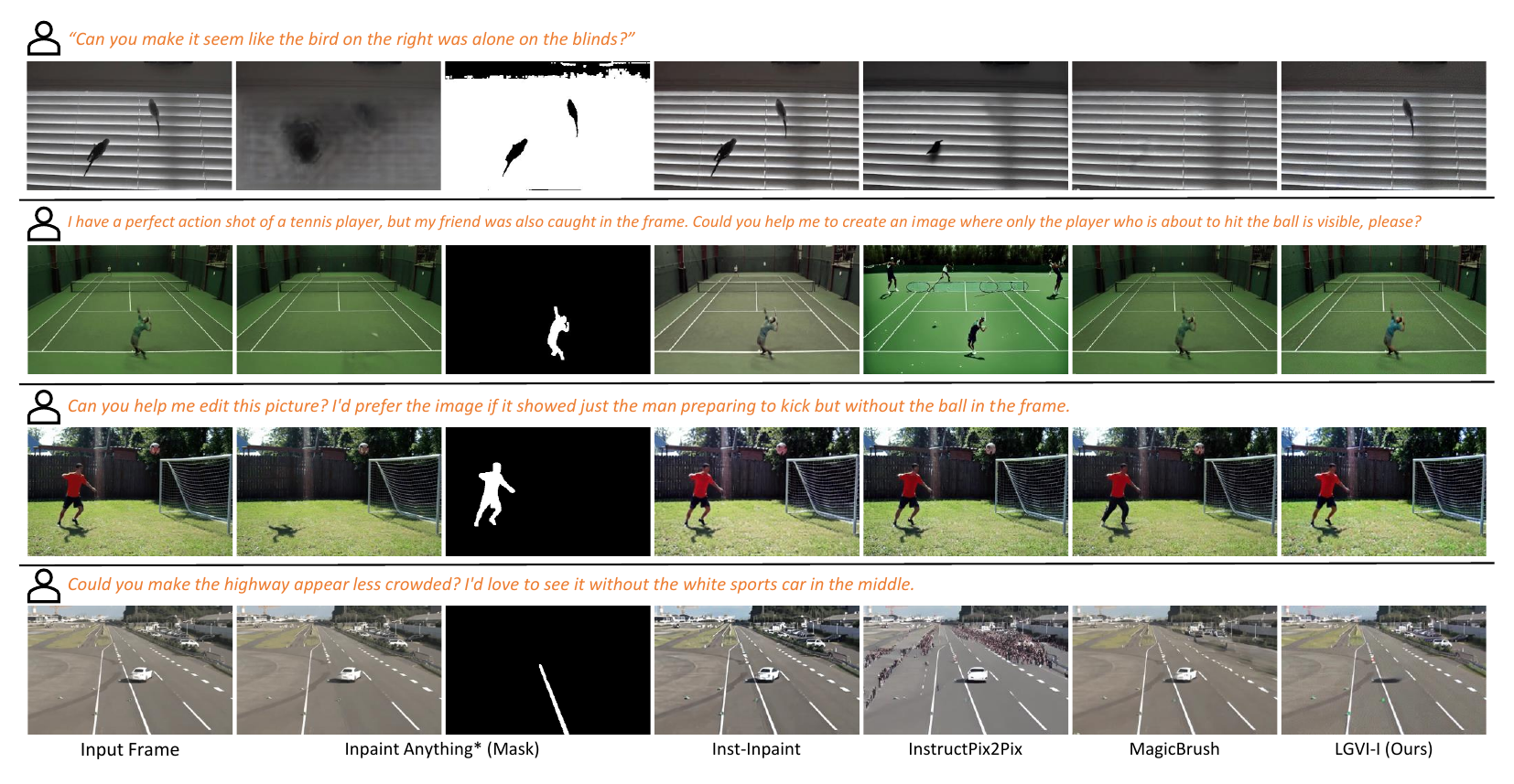}
	\caption{\small \textbf{Qualitative comparison between LGVI-I and baseline models on the interactive video inpainting task}. The chat-style conversation inputs are listed above each row. Columns 2 and 3 are the results and predicted masks from Inpaint Anything*. It removes the inaccurate objects according to the wrongly predicted masks due to the difficult interactive language inputs.}
	\label{fig:qualitative-comparison-interactive}
    \vspace{-5mm}
\end{figure*}

\section{Experiments}
\label{sec:exp}

\subsection{Settings}

\noindent
\textbf{Datasets and metrics.}
We use the ROVI dataset test set for both the referring video inpainting and interactive video inpainting tasks. The test set contains 478 videos and 758 objects, each equipped with one referring expression and one interactive expression. During the training of our models, we incorporate a referring image inpainting dataset, GQA-Inpaint~\cite{inst-inpaint}, to enrich the data vocabulary. We follow video inpainting works~\cite{liu2021fuseformer, video-inpainting-end-to-end, propainter, video-inpainting-flow-guided} to use PSNR and SSIM~\cite{ssim} to assess the statistical similarity between predicted results and ground truth. Additionally, we use VIFD~\cite{vfid} to measure the perceptual similarities. To assess the temporal consistency and smoothness of the generated videos, we also apply the $E_{warp}$ metric~\cite{Ewarp}.

\noindent
\textbf{Baselines.}
For the baselines, we select three language-driven image editing methods: InstructPix2Pix~\cite{instructpix2pix}, Inst-Inpaint~\cite{inst-inpaint}, and MagicBrush~\cite{magicbrush}. It is worth noting that InstructPix2Pix and MagicBrush are pre-trained on extensive image editing datasets. Inst-Inpaint is proposed to perform referring image inpainting on images. We also compare with Inpaint Anything, a multi-stage method for one-click video inpainting. It uses Segment Anything~\cite{segment-anything} and OSTrack~\cite{ostrack} to produce segmentation masks based on user click, followed by inpainting the masked area using inpainting models~\cite{STTN}. We implement Inpaint Anything*, which facilities Inpaint Anything~\cite{inpaint-anything} with GroundingDINO~\cite{groundingdino}, enabling it to process language inputs.

\noindent
\textbf{Implementation details.}
We initialize the U-Net weights from MagicBrush~\cite{magicbrush}. The newly introduced modules are trained from scratch. During training, we sample video and image data at a ratio of $3:1$. For the MLLM, we adopt LLaVA-7B~\cite{llava}. The learning rates are 3e-5, 1e-4, and 1e-4 for U-Net, mask decoder, and tuned parameters in MLLM, respectively. The loss weights are set to $\lambda_1=1$, $\lambda_2=0.001$, $\lambda_3=0.1$. These weights differ significantly due to the different types of losses they represent. The input and output video sizes are set to 512 $\times$ 320, and the video length is 24. For LGVI, we train 50 epochs on the ROVI dataset with a batch size of 32 for videos and 768 for images. For LGVI-I, we load the LGVI checkpoint and fine-tune it for 50 epochs under the same batch size. All experiments are carried out on 8 NVIDIA A100 GPUs.

\subsection{Referring Video Inpainting}

\noindent
\textbf{Quatitative results.}
We report quantitative results on the referring video inpainting task. Compared with baseline models, our model is the first one-stage language-driven video inpainting model. As shown in~\cref{tab:exp-refer-exprl}, our model outperforms MagicBrush~\cite{magicbrush} in all metrics and achieves on-par results with Inpaint Anything*~\cite{inpaint-anything}, even if Inpaint Anything* uses a mask-based inpainting model~\cite{STTN}. The results demonstrate the effectiveness of our model.

\noindent
\textbf{Qualitative results.}
Figure \ref{fig:qualitative-comparison-referring} shows qualitative results. We compare with MagicBrush~\cite{magicbrush}, a robust generalized language-driven image editing model. In the first example, where the language refers to the cat on the right, the MagicBrush model removes all the cats in the scene, while our model successfully inpaints the right cat. In the second example, the referring expression becomes more complex. MagicBrush struggles to identify the object requiring inpainting and removes the wrong object (the balls) in the last frame. In contrast, our model generates a plausible output, demonstrating its superior performance in handling complex language-driven inpainting tasks. Furthermore, in~\cref{fig:qualitative-multiple-nonexist}, we compare with Inpaint Anything* on sentences referring to multiple objects or nonexistent objects. Inpaint Anything is driven by a simple combination of referring segmentation and video inpainting models. Thus, it is fixed to produce one mask for each sentence. When referring to multiple or nonexistent objects, it outputs inaccurate results, while our model produces the correct output. This demonstrates the robustness of the language-driven setting.

\subsection{Interactive Video Inpainting}

\noindent
\textbf{Quatitative results.}
We report the interactive video inpainting task results in~\cref{tab:exp-chat-exprl}. As shown in the top 5 rows, when the models are trained using referring expressions, their performance drops correspondingly in this task. This is intuitive because interactive expressions are much longer and more implicit. For the MLLM-Enhanced Two-Stage Models, we combine the baseline models with an MLLM in a zero-shot manner. The interactive inputs are transferred into shorter referring expressions by simply prompting the MLLM. These models exhibit improved performances. Our LGVI-I model achieves the highest performance, demonstrating the effectiveness of the proposed architecture.

\noindent
\textbf{Qualitative results.}
Figure \ref{fig:qualitative-comparison-interactive} presents examples of the interactive video inpainting task. The user requests pose a significant challenge and complexity for the baseline models to comprehend. In particular, Inpaint Anything* predicts incorrect masks, leading to inaccurate results. Similarly, other diffusion-based models struggle to interpret the users' intentions accurately, resulting in less satisfactory outcomes. In contrast, our LGVI-I model, which harnesses the power of MLLM, consistently delivers the best performance in these challenging scenarios. This underscores the superiority of our proposed approach. More detailed ablations can be seen in the supplementary due to the page limitation.
\section{Conclusion}

In this paper, we propose a novel language-driven video inpainting task that uses language to guide inpainting areas. For training and evaluation, we collect a video dataset, namely ROVI. Comprehensive statistics demonstrate the uniqueness and diversity of our dataset, especially the chat-style interactive conversations generated by powerful LLMs and MLLMs. We further propose a diffusion-based baseline model, LGVI. Quantitative and qualitative experimental results show the effectiveness and robustness of our model. We hope our proposed benchmark and baselines can provide valuable insights into multi-modal models of low-level vision. In addition, there are several challenges to solve, including scalability and generalization of the model. We list the more discussion in the appendix.

\noindent
\textbf{Acknowledgement.} This project is supported by the National Key Research and Development Program of China (No.2023YFC3807600) and the National Key R\&D Program of China (No.2022ZD0161600). This study is also partially supported by the RIE2020 Industry Alignment Fund Industry Collaboration Projects (IAF-ICP) Funding Initiative, as well as cash and in-kind contribution from the industry partner(s).


\appendix

\noindent
\textbf{Overview.} Our supplementary includes the following sections:
\begin{itemize}
    \item \textbf{\cref{sec:supp-dataset-ann-details}.} Details for our dataset annotation process.
    \item \textbf{\cref{sec:baseline-details}.} Implementation details of the baseline models.
    \item \textbf{\cref{sec:supp-ablations}.} Quantitative ablation study results.
    \item \textbf{\cref{sec:supp-qualitative-results}.} More qualitative results of different models.
    \item \textbf{\cref{sec:supp-basics-diffusion}.} The foundations of Latent Diffusion Models and correlation with our model.
    \item \textbf{\cref{sec:supp-discussions}.} Discussions of limitations and challenges.
\end{itemize}

\noindent
\textbf{Video Demo.} We also include the video introduction of our work, which shows the visualization demo.

\section{Dataset Annotation Details}
\label{sec:supp-dataset-ann-details}

\begin{figure}[t!]
	\centering
	\includegraphics[width=1.0\linewidth]{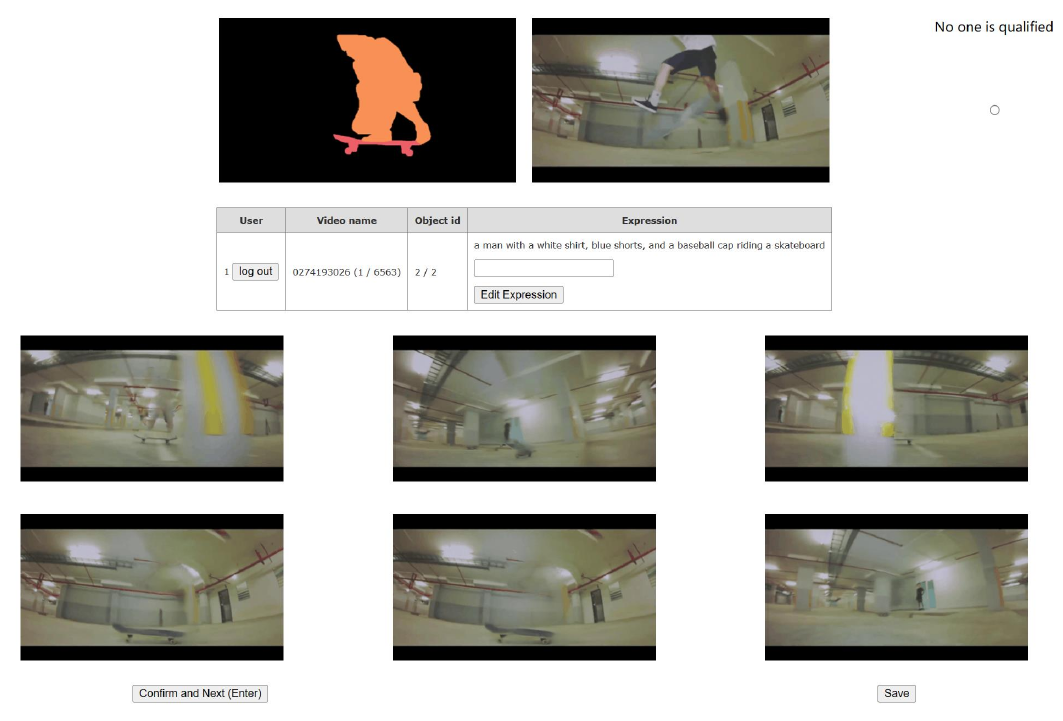}
	\caption{\small \textbf{The human annotation interface.} Human annotators click the best among the six results to select the results or choose ``No one is qualified'' to drop them all. The dialogue box in the middle is used to correct mis-annotated expressions.}
	\label{fig:supp-ann-helper}
\end{figure}

\subsection{Human Annotation}

When generating inpainting results using a video inpainting network~\cite{video-inpainting-end-to-end}, the input mask can be trickily expanded with different pixel sizes, denoted as $d$. The bigger $d$ is, the larger the input mask is developed so that it may cover the whole object. The best $d$ value varies through objects, causing an unstable performance in the inpainted videos if set to a fixed value.
Therefore, throughout the generation process, we experiment with various hyperparameters to generate multiple results for each object and involve human annotators to select the best result. 
In particular, we generate six samples with $d \in [0,3,5,7,10,15]$ for each object. Human annotators are expected to choose the best-looking result in these examples. The object does not enter the ROVI dataset if all examples are evaluated as unqualified. This human labeling process guarantees the high-quality ground truth of the ROVI dataset. \cref{fig:supp-ann-helper} shows an illustration of the human annotation interface.

Additionally, we find there are several \textit{misannotations} in the Refer-YouTube-VOS dataset. In some videos, the object label matches wrongly with object masks and expressions. For example, a person may correspond to ``a dog walking by a person'' according to the object labels. 
So, human annotators also make necessary revisions to the expressions in case of misannotations.

\begin{figure}[t!]
	\centering
	\includegraphics[width=1.0\linewidth]{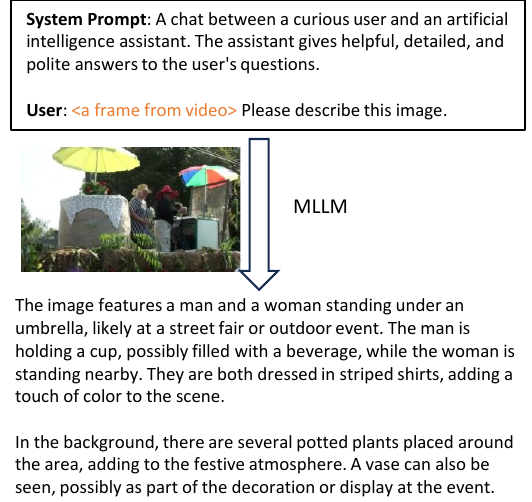}
	\caption{\small \textbf{The MLLM prompts in generating image descriptions.} System prompt here is the default system prompt of LLaVA~\cite{llava}}
	\label{fig:supp-interactive-pipeline-1}
\end{figure}

\begin{figure}[t!]
	\centering
	\includegraphics[width=1.0\linewidth]{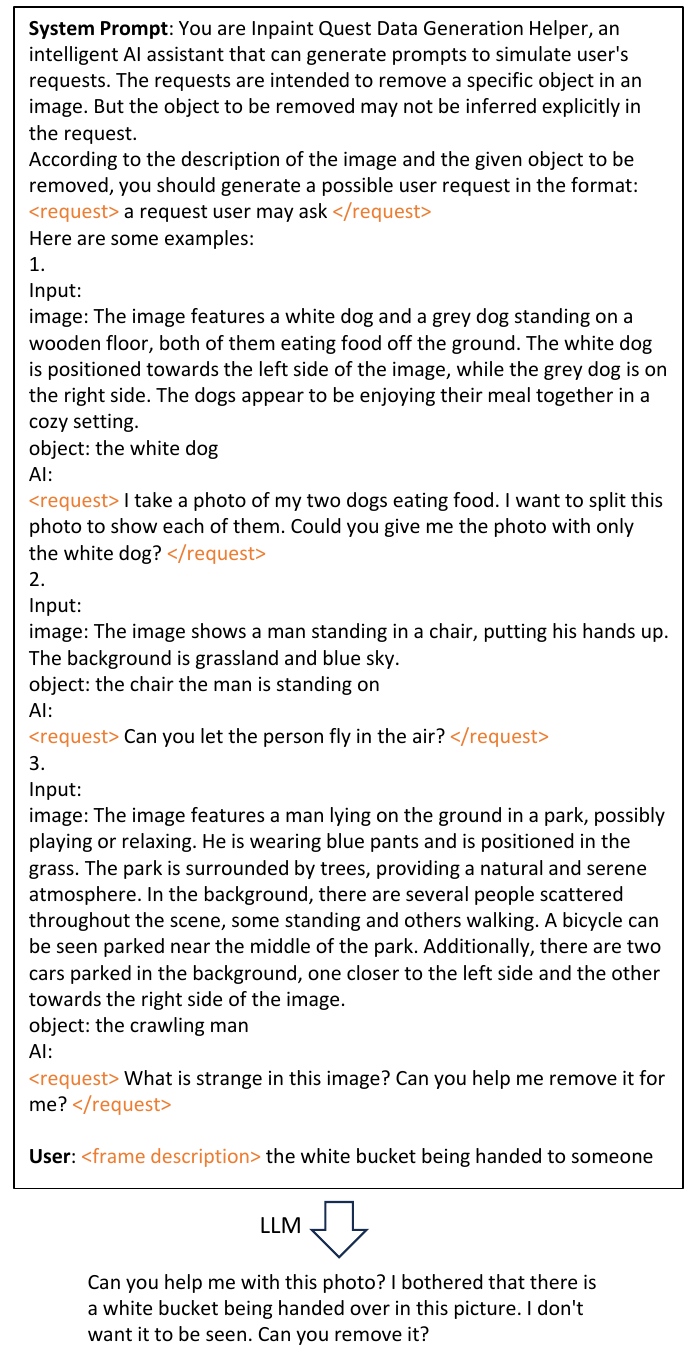}
	\caption{\small \textbf{The LLM prompts in generating use requests.}}
	\label{fig:supp-interactive-pipeline-2}
\end{figure}

\begin{figure}[t!]
	\centering
    \vspace{-1mm}
	\includegraphics[width=1.0\linewidth]{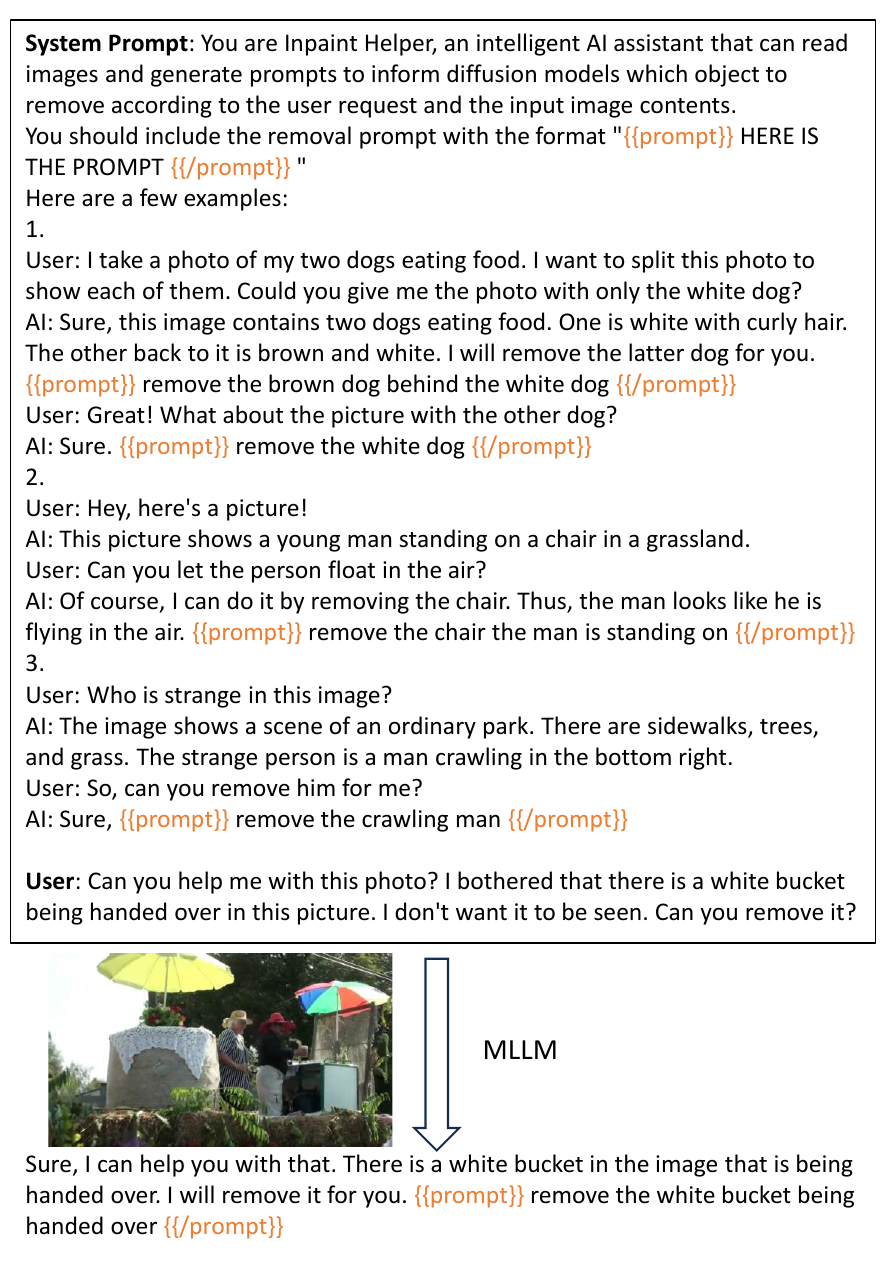}
	\caption{\small \textbf{The MLLM prompts in generating AI responses.}}
	\label{fig:supp-interactive-pipeline-3}
\end{figure}

\subsection{Interactive Annotation Details}

In the interactive annotation pipeline, all the generating processes are completed by prompting MLLMs and LLMs without fine-tuning. Firstly, we let an MLLM model generate a detailed description of a given frame. Then, we give the descriptions to an LLM and let it generate a possible user request. Finally, the MLLM generates AI responses according to the request and frame content. The pipeline is shown in~\cref{fig:supp-interactive-pipeline-1,fig:supp-interactive-pipeline-2,fig:supp-interactive-pipeline-3}.

\section{Baseline Details}
\label{sec:baseline-details}

For the Inst-Inpaint~\cite{inst-inpaint} baseline, we fine-tune the released checkpoint on the ROVI dataset with the same hyperparameters of LGVI. Specifically, we train 50 epochs on 8 × 80GB NVIDIA A100 GPUs with video and image batch sizes of 32 and 768. We resize the input image to 512 × 320, keeping the same as LGVI. For InstructPix2Pix~\cite{instructpix2pix}, MagicBrush~\cite{magicbrush}, and Inpaint Anything*~\cite{inpaint-anything}, we directly use their released checkpoints due to they are pre-trained on large scale image editing/inpainting datasets.

\begin{figure}[t!]
	\centering
	\includegraphics[width=0.95\linewidth]{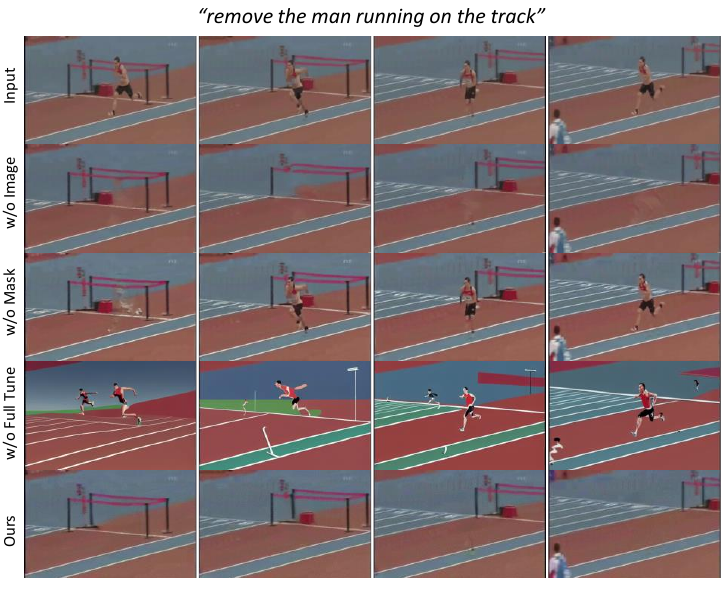}
	\caption{\small \textbf{Ablations.} We ablate makes supervision, image joint training, and fine-tuning the whole U-Net. Our model produces results with the highest vision and language coherence.}
	\label{fig:ablation-study}
    \vspace{-5mm}
\end{figure}

\begin{table}[!t]
    \centering
    \caption{\small \textbf{Quantitative ablations} on U-Net video inflation (VF), mask supervision (MS), image joint training (IJ), and fine-tuning the whole U-Net (FW). $E^*_{warp}$ denotes $E_{warp}(\times 10^{-2})$.}
    \label{tab:ablation-quantitative}
    \scalebox{0.98}{
    \hspace{-1.6mm}
    \begin{tabular}{c|cccc}
\toprule[0.15em]
Method & PSNR $\uparrow$ & SSIM $\uparrow$ & VFID $\downarrow$ & $E^*_{warp}$ $\downarrow$ \\
\hline
w/o FW & 20.53 & 0.607 & 0.370 & 1.101 \\
w/o MS & 21.80 & 0.631 & 0.358 & 1.059 \\
w/o VF & 22.08 & 0.754 & 0.356 & 1.017 \\
w/o IJ & 22.39 & 0.728 & \textbf{0.297} & 0.987 \\
\hline
LGVI (Ours) & \textbf{22.85} & \textbf{0.756} & 0.308 & \textbf{0.901} \\
\bottomrule[0.15em]
     \end{tabular}}
\end{table}

\section{Ablation Studies}
\label{sec:supp-ablations}

We conduct three ablations, including mask supervision, fine-tuning the entire U-Net, and joint training with images.

\noindent
\textbf{The benefit of mask supervision.}
The models without mask supervision rely solely on the inpainting ground for guidance, lacking an explicit signal to direct the inpainting area. As shown in~\cref{fig:ablation-study}, the running man remains present in all frames. Notably, although we utilize mask annotation in the ROVI dataset for training, the LGVI framework does not need mask input during inference. 

\noindent
\textbf{The benefit of fine-tuning the whole U-Net.}
We fine-tune the whole U-Net instead of only adjusting the parameters of new modules. As shown in~\cref{fig:ablation-study}, limiting the fine-tuning to only the new modules hinders training convergence, and the model struggles to output expected results. This demonstrates the necessity of tuning the whole model.

\noindent
\textbf{The benefit of joint training with images.}
As shown in~\cref{fig:ablation-study}, the model without joint training produces outputs with heavier artifacts than ours. This is because enlarging the dataset brings more diversity of objects and scenes to the model. It demonstrates the effectiveness of joint training.

\noindent
\textbf{Quantitative ablation studies.}
We show the quantitative ablation results in~\cref{tab:ablation-quantitative}. The observation is consistent with the ablations in the main paper. The performance without joint training with images drops slightly, except for a slight increase of 0.011 in the VFID metric. This can be attributed to the enhanced diversity of visual sources provided by additional image data. while it brings a compromise between the quality of the results and temporal consistency. The results also demonstrate the necessity of the U-Net inflation and mask supervision modifications of LGVI. The absence of these modifications leads to a noticeable reduction in performance. The most significant performance degradation is observed when the U-Net is not fully fine-tuned, but only the newly added parameters are trained. This decline can be attributed to the intrinsic differences between the inpainting task and the pre-trained image generation task. In the latter, the language input guides the model on what to create, but it does not specify what needs to be removed.

\begin{figure*}[t!]
	\centering
	\includegraphics[width=0.95\linewidth]{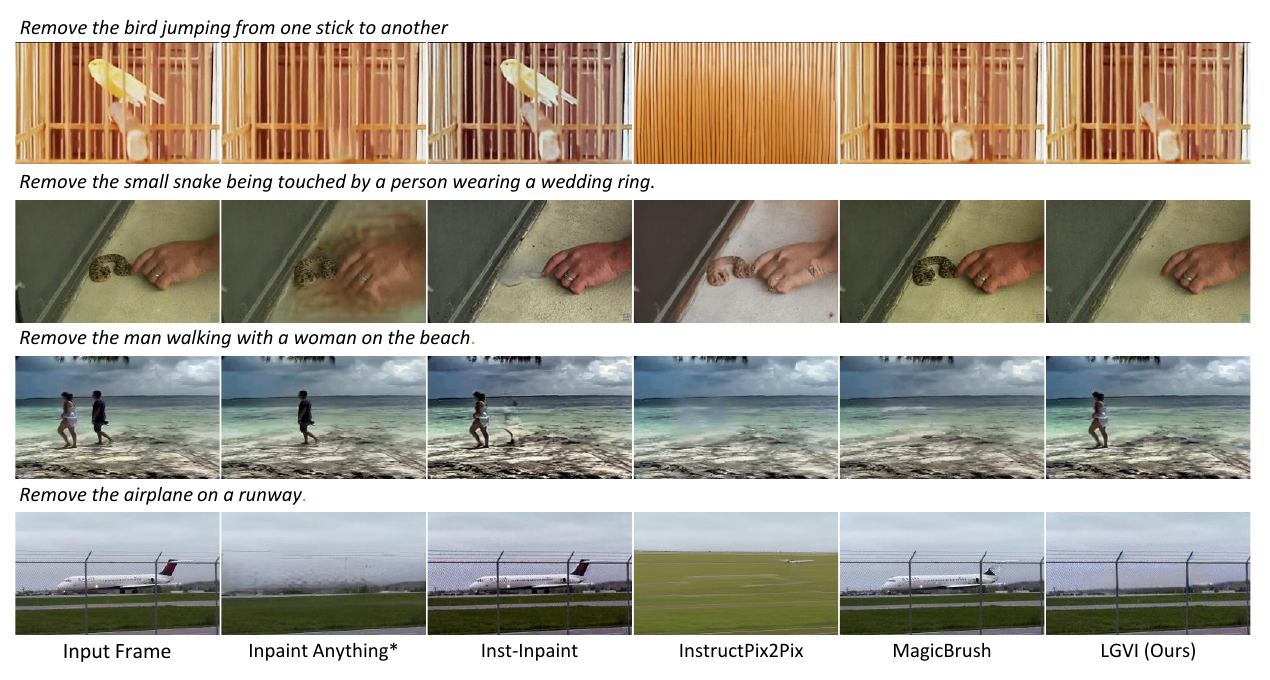}
	\caption{\small \textbf{More qualitative results for the referring video inpainting task.}}
	\label{fig:supp-qualitative-comparison}
\end{figure*}

\begin{figure*}[t!]
	\centering
	\includegraphics[width=0.95\linewidth]{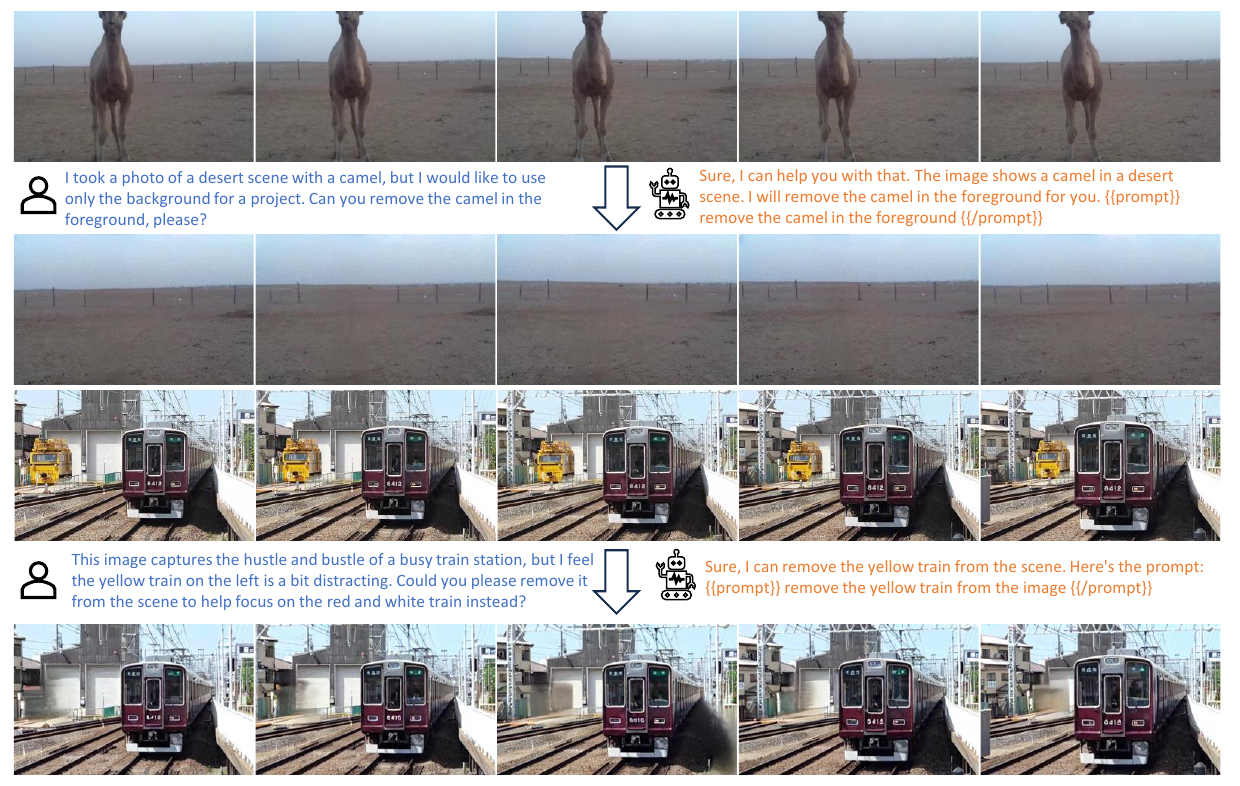}
	\caption{\small \textbf{More qualitative results for the interactive video inpainting task.}}
	\label{fig:supp-qualitative-interactive}
\end{figure*}

\section{More Qualitative Results}
\label{sec:supp-qualitative-results}

\cref{fig:supp-qualitative-comparison} compares the referring video inpainting task. It demonstrates the effectiveness of the proposed LGVI model. \cref{fig:supp-qualitative-interactive} shows the qualitative results of the interactive video inpainting task. Our LGVI-I model outputs both inpainting results and comprehensive text responses.

\begin{figure*}[t!]
	\centering
	\includegraphics[width=1.0\linewidth]{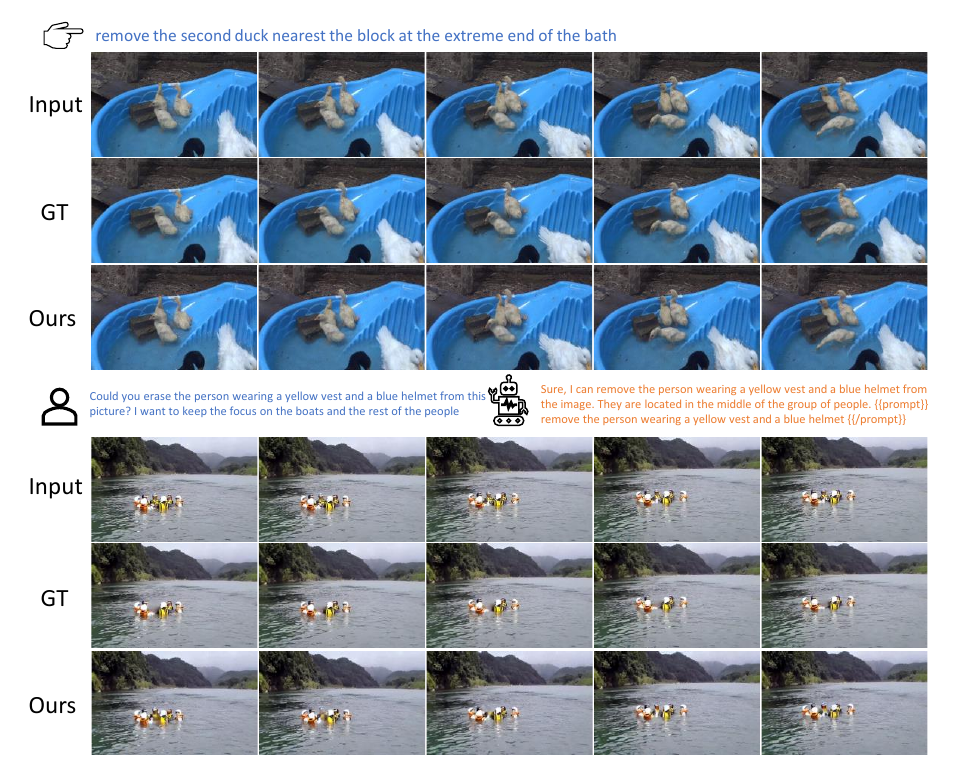}
	\caption{\small \textbf{Failure cases.} In the first example, where the referring expression is vague and implicit, our model fails at removing the right duck. In the second example, where many instances in the same category occur together, our model can hardly recognize the user-intended one. However, the language output describes the approximate position of the referred person.}
	\label{fig:supp-limitation}
\end{figure*}

\section{Basics of Diffusion Models}
\label{sec:supp-basics-diffusion}

\noindent
\textbf{Denoising Diffusion Probabilistic Models (DDPMs).}
The core of DDPMs~\cite{ddpm} involves iteratively adding noise to the data until it becomes a sample from a simple Gaussian distribution. The reverse process, which generates data from the noise, is learned by the model. The forward process, also known as the ``noising'' process, is typically modeled as a Markov chain that gradually adds Gaussian noise to the data over a sequence of time steps $T$:
\begin{equation}
    x_0 \sim q(x_0), x_{t} = \sqrt{\alpha_{t}} x_{t-1} + \sqrt{1 - \alpha_{t}} \epsilon,
\end{equation}
where $x_0$ is a sample from the data distribution $q(x_0)$, $x_t$ represents the data at time step $t$, and $\alpha_{t}$ is a variance schedule that determines the amount of noise to add at each step. $\epsilon$ is the noise sampled from a standard Gaussian distribution $\mathcal{N}(0, \mathbf{I})$. The reverse process, which is the generative process, aims to learn the distribution of the original data by reversing the noising process. This involves learning a parameterized function $\theta$ that models the reverse conditional probability $p_{\theta}(x_{t-1}|x_{t})$. The reverse process is described by:
\begin{equation}
    p_{\theta}(x_{t-1}|x_{t}) = \mathcal{N}(x_{t-1}; \mu_{\theta}(x_{t}, t), \Sigma_{\theta}(x_{t}, t)),
\end{equation}
where $\mu_{\theta}(x_{t}, t)$ and $\Sigma_{\theta}(x_{t}, t)$ are learned functions that predict the mean and covariance of the distribution for $x_{t-1}$, given $x_t$ and time step $t$.
The learning of $\theta$ is typically done via a variational approach, minimizing a loss function that is a modified version of the Evidence Lower BOund (ELBO). This loss function ensures that the learned reverse process closely approximates the true distribution of the data, which can be simplified as:
%
\begin{equation}
    L(\theta) = \mathbb{E}_{x, \epsilon \sim \mathcal{N}(0, 1), t} \left [ || \epsilon - \epsilon_\theta(x_t, t) ||^2_2 \right ].
\end{equation}
The denoising process can incorporate extra guidance, where the model is trained to generate samples conditioned on a set of labels or attributes $c$. Typical guidances are language and images~\cite{stable-diffusion, controlnet}. The loss function can be updated as follows:
\begin{equation}
    L(\theta) = \mathbb{E}_{x, \epsilon \sim \mathcal{N}(0, 1), t, c} \left [ || \epsilon - \epsilon_\theta(x_t, t, c) ||^2_2 \right ],
\end{equation}
where $c$ is guidance to control the generation result. We extend the condition input with a video input $\mathbf{X}$ to control the inpainting results.

\noindent
\textbf{Latent Diffusion Models (LDMs).}
Latent Diffusion Model (LDM)~\cite{stable-diffusion} is a type of generative model that operates on a latent space rather than directly on the data space. The primary idea is to first encode high-dimensional data, like images, into a lower-dimensional latent representation $z = \mathcal{E}(x)$ and then apply the diffusion process within this latent space. A decoder reconstructs the latent back to the pixel distribution $x = \mathcal{D}(z)$. 

\noindent
\textbf{LGVI Architecture.}
The core of our model is to produce inpainting results $\hat{\mathbf{Y}}$ driven by language guidance $c$ and vision input $\mathbf{X}$. This core concept is versatile and can be integrated into a variety of existing architectures, including those based on diffusion or transformer paradigms, as long as the model can fuse language and vision inputs. We choose the LDM architecture because of its flexibility in inflating to video modality and improved sample quality due to the reduced dimensionality of the problem.

\section{Future Work Discussions}
\label{sec:supp-discussions}

\noindent
\textbf{Failure cases.}
\cref{fig:supp-limitation} shows two LGVI and LGVI-I failure cases. The models still face the core challenge of implicit language or description. In the first example, the referring expression is comparatively long and hard to understand, leading to poor performance. In the second example for the interactive task, even if the MLLM outputs a reasonable response and correctly predicts the position of the removed person, the diffusion model does not provide the right output. That is because the current diffusion-based baseline model is a preliminary modification of a pre-trained text-to-image model, which lacks the understanding of precise locations. The failure cases demonstrate that despite the comparatively stronger performance of previous methods, our proposed model is still a baseline in the language-driven video inpainting field. Future work is expected to develop more advanced methods to overcome the challenges.

\noindent
\textbf{Challenges.}
(1) \textbf{Handling Ambiguity in Language Descriptions.} Language-driven video inpainting relies heavily on the accuracy and clarity of language inputs. Ambiguities or vagueness in language descriptions can lead to inaccuracies in inpainting results. Developing models that can intelligently handle or clarify ambiguous language inputs is a significant challenge. (2) \textbf{Real-Time Processing.} Video inpainting in a real-time setting, especially with complex language-driven inputs, is computationally demanding. Diffusion-based models also experience the slow inference problem due to the Markov denoising process. Improving the speed and efficiency of these models without compromising accuracy is a crucial challenge. (3) \textbf{Scalability and Generalization.} Another challenge is ensuring that the model generalizes well across various languages and video types. Models might perform well on the dataset they were trained on but struggle with new, unseen data.

\noindent
\textbf{Future work.}
A promising direction is to resolve ambiguities in language inputs, possibly by using contextual clues from the video or previous language inputs. In addition, researching methods to optimize these models for real-time video inpainting, could be valuable for live broadcasting or interactive media. Another important future work is incorporating interactive user feedback mechanisms that allow the system to learn from corrections or preferences indicated by users, thereby improving the accuracy and relevance of the inpainting results over time. See supplementary for more discussions.

\noindent
\textbf{Potential social impacts.}
This technology could potentially boost creative fields such as film-making, advertising, and content creation. It allows for more seamless editing and creative storytelling, enabling creators to modify and enhance their visual narratives easily. It also comes with negative impacts. For example, it can be used in creating misleading or false media and ethical or moral issues.

{
    \small
    \bibliographystyle{ieee_fullname}
    \bibliography{egbib}
}

\end{document}